\documentclass{article}

\usepackage[final]{neurips_2022}
\usepackage{multirow}
\usepackage{graphicx}
\usepackage{subcaption}
\usepackage{amsmath}

\setcitestyle{numbers,square}

\usepackage[utf8]{inputenc} 
\usepackage[T1]{fontenc}    
\usepackage{hyperref}       
\usepackage{url}            
\usepackage{booktabs}       
\usepackage{amsfonts}       
\usepackage{nicefrac}       
\usepackage{microtype}      
\usepackage{xcolor}         

\usepackage{amsmath}
\DeclareMathOperator*{\argmax}{arg\,max}

\title{Calibrating Where It Matters: \\Constrained Temperature Scaling}

\author{%
  Stephen J.~McKenna
  \\
  School of Science and Engineering\\
  University of Dundee\\
  Dundee DD1 4HN, UK \\
  \texttt{s.j.z.mckenna@dundee.ac.uk} \\
  \And
  Jacob Carse \\
  School of Science and Engineering \\
  University of Dundee \\
  Dundee DD1 4HN, UK \\
  \texttt{j.carse@dundee.ac.uk} \\
}

\begin{document}

\maketitle

\begin{abstract}
We consider calibration of convolutional classifiers for diagnostic decision making. Clinical decision makers can use calibrated classifiers to minimise expected costs given their own cost function. Such functions are usually unknown at training time. If minimising expected costs is the primary aim, algorithms should focus on tuning calibration in regions of probability simplex likely to effect decisions. We give an example, modifying temperature scaling calibration, and  demonstrate improved calibration where it matters using convnets trained to classify dermoscopy images. 
\end{abstract}

\section{Classifier Calibration for Medical Decision Making}

When using an image classifier for medical diagnostic decision support, it is important that the classifier quantifies the uncertainty of (or confidence in) its predictions. Obtaining well calibrated class predictions is important in this regard. A classifier that predicts class probabilities is said to be {\em distribution calibrated} if, for those test images assigned a predicted probability vector $\mathbf{p}$, the classes are distributed approximately as $\mathbf{p}$. In multi-class settings, weaker notions of calibration are adopted, such as {\em confidence calibration} in which only the probability for the class with the largest predicted probability is calibrated~\citep{Filho2023MachineLearning,Guo2017Calibration,Zhao2021Neurips}.  
Deep classifiers often provide predictions that are not well-calibrated~\citep{Guo2017Calibration} although the quality of calibration varies depending on factors such as neural architecture~\citep{Minderer2021Revisiting}. This has led to increased interest in algorithms for improving calibration using methods for regularising training, post-hoc calibration of trained networks, or their combinations~\citep{Wang2021RethinkingCalibration}.

Calibrated classifiers can be used to decide upon actions (e.g., whether or not to refer a patient for further investigation) in a way that minimises expected cost~\citep{Zhao2021Neurips,Filho2023MachineLearning}. In binary  settings, the relative costs of false positive and false negative errors (along with class priors) yield a decision threshold that will minimise expected cost. In the multi-class setting, costs incurred by confusing disease classes can be specified as a cost function given in the form of a cost matrix. Cost functions will differ between the healthcare settings in which a trained classifier might be deployed. They will also differ from the loss function used to train the classifier. However, we wish to emphasise that although these cost functions are unlikely to be known at training time, they are nevertheless constrained by prior knowledge. 

We argue that it is not important to obtain equally well calibrated predictions over the entire probability simplex. This is because the set of cost functions that might be used at deployment is constrained. Calibration is most important in regions where decision boundaries might lie. It is not necessary to calibrate as if all possible cost functions might be used. For example, when developing a classifier for a triage setting we might be certain {\em a priori} that the costs of misclassifying maligant cases as benign will be greater than the costs of misclassifying benign cases as malignant. This constrains the region in which decision boundaries will lie, in which obtaining calibration is most important. We suggest that calibration algorithms might be developed with such constraints in mind, focusing on calibrating where it matters most. 
In this short paper, we consider temperature scaling, a popular single-parameter, post-hoc calibration method. By modifying parameter estimation, we obtain classifiers with calibration tailored to clinically motivated constraints on the cost functions. We demonstrate this using convolutional neural classifiers trained for binary and multi-class classification on the ISIC 2019 challenge dermoscopy dataset.

\section{Estimating Temperature for Post-hoc Calibration}
\label{sec:TS}





Temperature scaling is a post-hoc method used to improve calibration of deep classifiers~\citep{Guo2017Calibration}.
It scales the logits, $z$, using a single temperature parameter $T$ that is estimated from a validation set (or "calibration set") by minimising negative log-likelihood. 

The first example we consider is one of binary classification of images as either malignant (positive) or benign (negative). A deep classifier computes the probability $p = \sigma(z/T)$ that an image is malignant by applying a logistic function to its output neuron's logit $z$ after scaling it by $1/T$. We would like the trained classifier to be calibrated so that it can be deployed with different cost functions. We don’t know what these functions will be exactly but we might know something about them. Suppose we know that the decision threshold (on $p$) will always be in the range $(0.0, 0.5)$. In this case, the classifier needs to be well calibrated when $p<0.5$. When $p>0.5$, any decision maker will decide to act as if the image is malignant so it would be sufficient if observed frequencies were at least $0.5$ in that upper range of $p$. Optimisation of temperature, $T$, can be modified to emphasise calibration near the possible decision boundaries by estimating negative log-likelihood from the subset of the calibration set consisting of those images with predictions $p<0.5$ (equivalently $z<0$). Note that changing the temperature has no effect on the membership of this subset. We denote temperature estimated in this way by $T^*$.


Second, consider multi-class classification in which some classes are benign and the others malignant. We again make an assumption about the cost functions that will be used at deployment: that the most costly mistakes involve misclassifying images from benign classes. The algorithm for estimating $T$ can be modified to emphasise calibration near the possible decision boundaries by estimating negative log-likelihood from the subset of the calibration set consisting of those images where the most probable class $\hat{y} = \argmax_k \frac{\exp(\frac{z_k}{T})}{\sum_j\exp(\frac{z_j}{T})}$ is a benign class. Again, temperature has no effect on membership of this subset, and we denote temperature estimated this way by $T^*$.


\section{Experiments}

We report experiments\footnote{GitHub repository: \url{https://github.com/UoD-CVIP/Calibration_Where_it_Matters}} on the ISIC 2019 challenge dataset of dermoscopic images of skin lesions~\citep{codella2018skin,combalia2019bcn20000,Tschandl2018Ham10000}. We used images from seven classes, three of which were benign (melanocytic nevus, benign keratosis, and dermatofibroma) and four malignant (melanoma, basal cell carcinoma, actinic keratosis, and squamous cell carcinoma). Data were split into fixed training, validation, and test sets in the proportions $60:20:20$. The validation set was used for early stopping and temperature estimation. We report results for two convolutional architectures previously used with this dataset: EfficientNet~B7~\citep{Tan2019EfficientNet} and ResNet101~\citep{He2016DeepResidual}, both pre-trained on ImageNet.

\begin{figure}[tb!]
  \centering
  \begin{subfigure}{0.5\textwidth}
    \includegraphics[width=\linewidth]{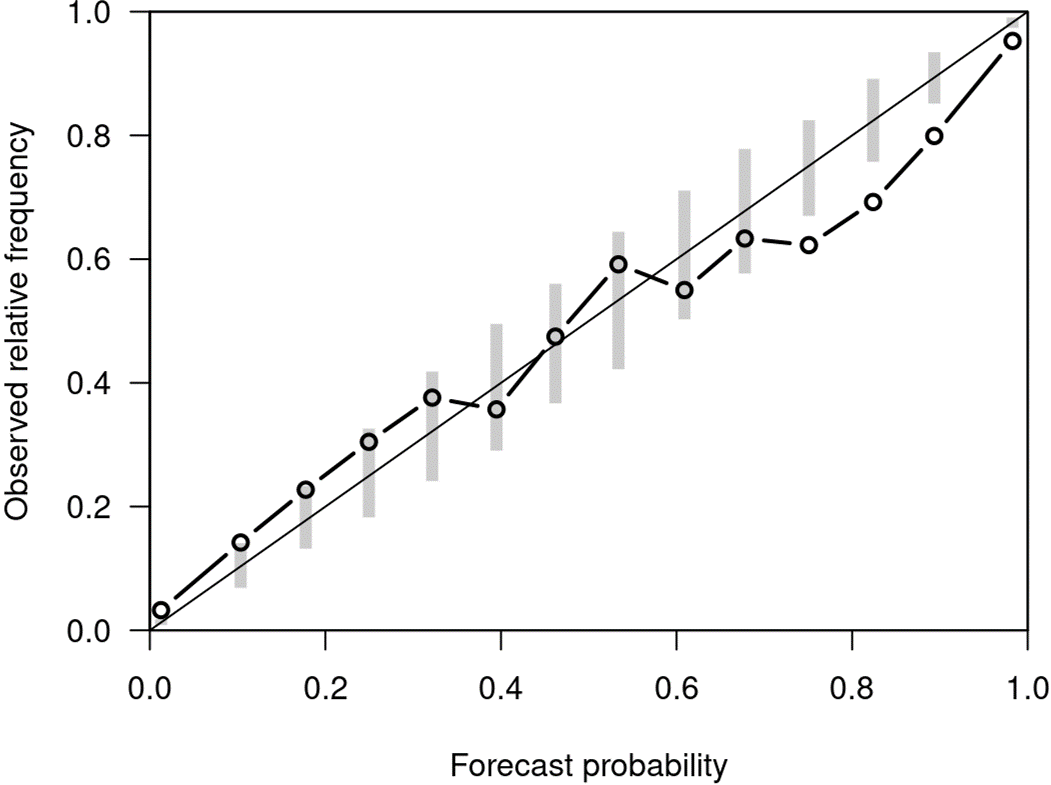}
  \end{subfigure}%
  \begin{subfigure}{0.5\textwidth}
    \includegraphics[width=\linewidth]{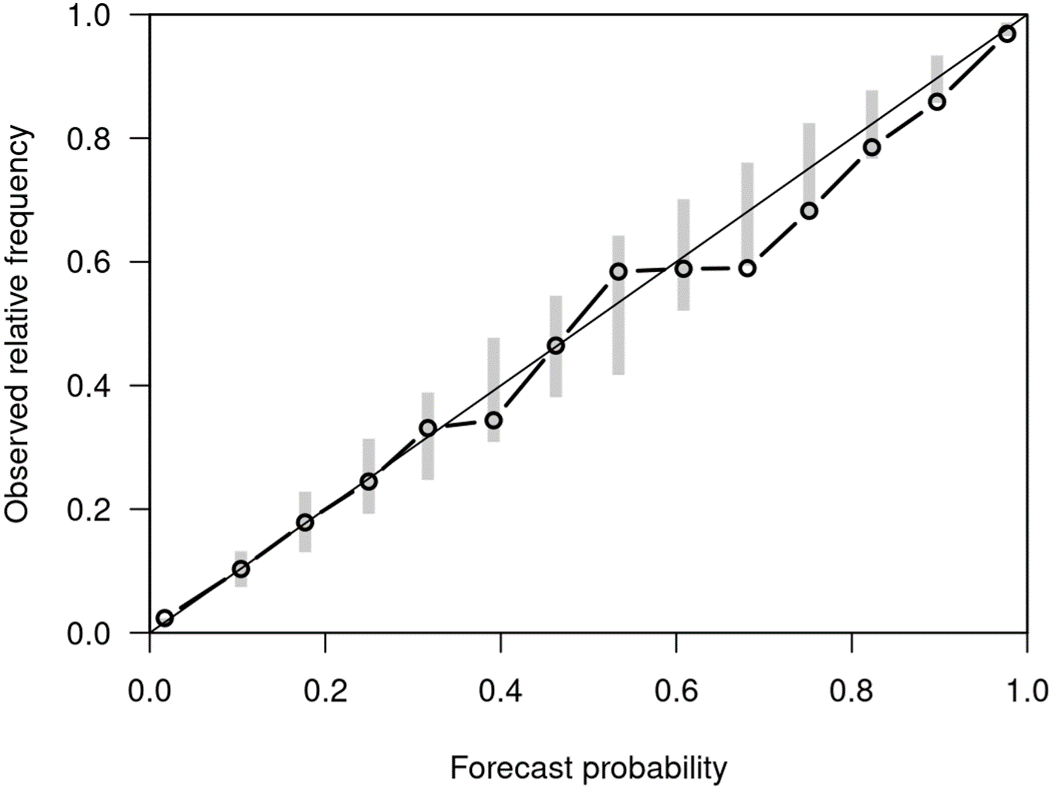}
  \end{subfigure}
  \caption{
  Reliability diagrams for binary EfficientNet~B7, without calibration (left) and with modified temperature scaling (right). Vertical bars indicate $5\%-95\%$ bootstrap consistency intervals~\citep{Brocker2007ReliabilityDiagrams}.}
  \label{fig:reliability_diagrams}
\end{figure}

Figure~\ref{fig:reliability_diagrams} shows reliability diagrams for EfficientNet~B7 binary classifiers trained to discriminate maligant from benign lesions. The classifier without post-hoc calibration is not well calibrated. In contrast, the result after scaling with $T^*$ estimated as described in Section~\ref{sec:TS} is consistent with reliability. Furthermore, that classifier is especially highly calibrated in the range $p\leq0.5$ as desired.

Table~\ref{tab:binary} reports estimated calibration errors (ECE) separately for test examples with predicted probability of malignant less than $0.5 (z<0)$ and otherwise $(z \geq 0)$. Standard temperature scaling reduced ECE in both ranges for both architectures. Using $T^*$ as the estimated temperature further reduced ECE in the range $p<0.5$ while retaining much of the benefit of temperature scaling in the range $p \geq 0.5$. 

Table~\ref{tab:multi_class} reports ECEs separately for test examples with maximum probability assigned to one of the benign classes ($ECE_{Ben}$) and to one of the maligant classes ($ECE_{Mal}$). We used class-wise ECE~\citep{Filho2023MachineLearning}. Again, standard temperature scaling reduced ECE in both regions for both architectures. Using $T^*$ as the estimated temperature further reduced $ECE_{Ben}$ (albeit marginally in the case of ResNet) but without increasing $ECE_{Mal}$. 


\begin{table}[!tb]
	\centering
	\caption{Binary classification results. Lowest ECE values are in bold. Bootstrap 95\% confidence intervals are given. AUC denotes area under the ROC curve (one vs. rest).}
	\label{tab:binary}
	{
		\begin{tabular}{crrlrr}
			\hline
			Architecture & \multicolumn{1}{c}{\begin{tabular}[c]{@{}c@{}}Balanced\\ Accuracy\end{tabular}} & \multicolumn{1}{c}{AUC} & \multicolumn{1}{c}{\begin{tabular}[c]{@{}c@{}}Calibration\\ Method\end{tabular}} & \multicolumn{1}{c}{$\textnormal{ECE}_{z<0}$} & \multicolumn{1}{c}{$\textnormal{ECE}_{z \geq 0}$} \\ \hline
			\multirow{3}{*}{\begin{tabular}[c]{@{}c@{}}EfficientNetB7\end{tabular}} & \multirow{3}{*}{88.0\%} & \multirow{3}{*}{0.944} & None & \(0.0264 \pm 0.0008\) & \(0.0525 \pm 0.0014\) \\
			&  &  & Temp Scaling (\(T\)) & \(0.0080 \pm 0.0007\) & \(\textbf{0.0135} \pm 0.0011\) \\
			&  &  & Temp Scaling (\(T^*\)) & \(\textbf{0.0058} \pm 0.0006\) & \(0.0287 \pm 0.0014\) \\ \hline
			\multirow{3}{*}{\begin{tabular}[c]{@{}c@{}}ResNet101\end{tabular}} & \multirow{3}{*}{86.3\%} & \multirow{3}{*}{0.934} & None & \(0.0295 \pm 0.0009\) & \(0.0383 \pm 0.0014\) \\
			&  &  & Temp Scaling (\(T\)) & \(0.0161 \pm 0.0008\) & \(\textbf{0.0156} \pm 0.0010\) \\
			&  &  & Temp Scaling (\(T^*\)) & \(\textbf{0.0151} \pm 0.0007\) & \(0.0167 \pm 0.0010\) \\ \hline
		\end{tabular}%
	}
\end{table}

\vspace*{-\baselineskip}
\begin{table}[!tb]
	\centering
	\caption{Results for multi-class classifiers.}
	\label{tab:multi_class}
	{
		\begin{tabular}{crrlrr}
			\hline
			Architecture & \multicolumn{1}{c}{\begin{tabular}[c]{@{}c@{}}Balanced\\ Accuracy\end{tabular}} & \multicolumn{1}{c}{AUC} & \multicolumn{1}{c}{\begin{tabular}[c]{@{}c@{}}Calibration\\ Method\end{tabular}} & \multicolumn{1}{c}{$\textnormal{ECE}_{Ben}$} & \multicolumn{1}{c}{$\textnormal{ECE}_{Mal}$} \\ \hline
			\multirow{3}{*}{\begin{tabular}[c]{@{}c@{}}EfficientNetB7\end{tabular}} & \multirow{3}{*}{75.1\%} & \multirow{3}{*}{0.974} & None & \(0.0156 \pm 0.0002\) & \(0.0258 \pm 0.0004\) \\
			&  &  & Temp Scaling (\(T\)) & \(0.0103 \pm 0.0002\) & \(0.0120 \pm 0.0002\) \\
			&  &  & Temp Scaling (\(T^*\)) & \(\textbf{0.0086} \pm 0.0002\) & \(\textbf{0.0117} \pm 0.0002\) \\ \hline
			\multirow{3}{*}{\begin{tabular}[c]{@{}c@{}}ResNet101\end{tabular}} & \multirow{3}{*}{70.8\%} & \multirow{3}{*}{0.967} & None & \(0.0160 \pm 0.0003\) & \(0.0266 \pm 0.0004\) \\
			&  &  & Temp Scaling (\(T\)) & \(0.0129 \pm 0.0002\) & \(0.0147 \pm 0.0003\) \\
			&  &  & Temp Scaling (\(T^*\)) & \(\textbf{0.0126} \pm 0.0002\) & \(\textbf{0.0145} \pm 0.0003\) \\ \hline
		\end{tabular}%
	}
\end{table}

\section{Discussion}

We have demonstrated how modifying estimation of the temperature scaling parameter can result in better calibration where it matters to inform decision making, i.e., in regions of probability simplex likely to be near decision boundaries. The method can be applied as described to other accuracy-preserving post-hoc calibration methods~\citep{Frenkel2022Calibration} or in combination with decision calibration~\citep{Zhao2021Neurips}. We have described an example of calibration that takes account of constraints on the space of cost functions (in combination with class priors). Exploring this more generally for other cost function priors and calibration methods is a topic for future work. 

A limitation is the potential to adversely effect estimation of the overall expected cost. If reporting confidences as well as decisions, probabilities can be reported in the well-calibrated region of the simplex, whereas elsewhere it might be more appropriate to report, e.g., that "malignant is more probable than benign". This preliminary study should be extended to multiple datasets and other architectures, and robustness to dataset shift investigated~\citep{Minderer2021Revisiting,Ovadia2019CanYouTrust}. 

\begin{ack}
This paper reports independent research funded by the National Institute for Health and Care Research (Artificial Intelligence, Deep learning for effective triaging of skin disease in the NHS, AI AWARD01901) and NHS Transformation Directorate. The views expressed in this publication are those of the authors and not necessarily those of the National Institute for Health and Care Research, NHS Transformation Directorate or the Department of Health and Social Care. This research also received funding from the Tayside Dermatological Research Charity.
\end{ack}

{
\small
\bibliographystyle{plainnat}
\bibliography{refs}
}

\end{document}